\documentclass[a4paper]{article}

\usepackage[final]{neurips_2024}

\usepackage[utf8]{inputenc} 
\usepackage[T1]{fontenc}    
\usepackage{hyperref}       
\usepackage{url}            
\usepackage{booktabs}       
\usepackage{amsfonts}       
\usepackage{nicefrac}       
\usepackage{microtype}      
\usepackage{xcolor}         

\usepackage{doi}
\usepackage{tcolorbox}

\def\Snospace~{\S{}}

\newcommand{\hreffoot}[2]{\href{#1}{#2}\footnote{\url{#1}}}

\definecolor{linkcolor}{HTML}{991408}  
\definecolor{citecolor}{HTML}{2E7E2A}  
\definecolor{filecolor}{HTML}{131877}  
\definecolor{menucolor}{HTML}{727500}  
\definecolor{runcolor} {HTML}{137776}  
\definecolor{urlcolor} {HTML}{0a2bbf}  
\hypersetup{colorlinks=true,linkcolor=linkcolor,citecolor=citecolor,filecolor=filecolor,menucolor=menucolor,runcolor=runcolor,urlcolor=urlcolor}

\newcommand{\hider}[1]{}

\title{System 2 Reasoning Capabilities Are Nigh}

\author{%
  Scott~C. Lowe \\
  Vector Institute \\
  Toronto, Canada \\
  \texttt{scott.lowe@vectorinstitute.ai} \\
}

\begin{document}

\maketitle

\begin{abstract}
In recent years, machine learning models have made strides towards human-like reasoning capabilities from several directions.
In this work, we review the current state of the literature and describe the remaining steps to achieve a neural model which can perform System~2 reasoning analogous to a human.
We argue that if current models are insufficient to be classed as performing reasoning, there remains very little additional progress needed to attain that goal.
\end{abstract}

\section{Introduction}

%
%
%
%
%

The dual process theory of thought processes is long standing within psychology \citep{Wason1974, Evans2008, Stanovich_West_2000} and was popularized more broadly by \citet{kahneman_thinking_2012}.
In this framework, human thinking capabilities are conceptualized as two distinct modes of thought.
System~1 is fast, automatic, instinctive, and more emotional; System~2 is slower, effortful, deliberate, and more logical.
System~1 is, in essence, unconscious thought, and System~2 is conscious thought; though there is not yet consensus on whether ``ah-ha'' moments which come following an incubation period are triggered by unconscious work of System~1 or 2 \citep{christensen2005problematic,incubation}.
Additionally, due to its instinctive and reactive nature, System~1 is more prone to bias than System~2, though System~2 is not without bias  \citep{Tversky1974}. 

In comparison to these two cognitive systems of thought, feed-forward neural networks are sometimes described as being analogous to System~1.
Their outputs are immediate and automatic, yielded immediately without what might call ``deliberation''.
Like with System~1, the computational system producing the output does not and can not provide an explicit explanation for \textit{why} it produced a certain response, making interpretability challenging, even when attempting to induce it to provide an a posteriori justification for its response \citep{jung-etal-2022-maieutic}.
Such systems are effectively performing pattern matching for their current stimulus against the body of data imbibed during training.

In comparison, symbolic rule-based algorithms (classical ``artificial intelligence''), whether they are manually or programatically created, can provide an explanation for their reasoning.
However their performance is limited because the space of the real-world is too large to be handled with a narrow set of rules that are coded in the stimulus domain \citep{yihe2019}.

In this work, we review the existing literature in the space of reasoning from the perspective of cognitive psychology and machine learning, and we speculate on what form a neural network would need to take for it to be able to perform reasoning in the style of System~2.
We argue the majority of hurdles needed to achieve this task have already been cleared, and there are a small number of pieces of the puzzle remaining.
Thus complex agents, trained through deep learning, that can reason logically about the real-world will be available in the near-term, if they are not here already.

\section{Background}

\subsection{Modalities of human thought}
\label{sec:bg:modes-of-thought}

Historic texts indicate that ancient philosophers such as Plato believed that thinking was synonymous with an inner monologue \citep{Wiley2006}.
However, whilst an internal monologue (inner speech) is common, it is not ubiquitous and most people overestimate how often their thoughts are expressed verbally \citep{Hurlburt2013}.
There is a wide variety of inner experiences across humans \citep{Hurlburt2006}, and most people are surprised when they first discover that other people's internal experiences differ greatly from their own.

The modalities of human inner thought are \citep{Hurlburt2011, Hurlburt2013}:
\begin{itemize}
    \item Inner speaking/inner monologue --- thoughts expressed verbally, e.g. talking to yourself, hearing your/a voice while recalling.
    \item Inner seeing/visual imagery --- thoughts expressed visually, e.g. picturing a memory or imagining a hypothetical scene.
    \item Feelings --- a conscious experience of emotional processes, e.g. sadness when grieving.
    \item Unsymbolized thinking --- thoughts expressed without words or images, e.g. drinking a glass of water, without internal discussion or commentary.
    \item Sensory awareness --- attending to a sensory aspect of the environment for an unimportant reason, e.g. hearing someone talk but seeing the light reflecting off their glasses.
\end{itemize}

Most people experience inner speech and inner imagery some of the time but not all of the time, with the majority of their thought processes unsymbolized \citep{Hurlburt2013}.
However there are outliers in each direction, with some people having no inner speech (anauralia), constant inner speech, no mind's eye (aphantasia), or extremely vivid mental imagery as detailed as sensory stimuli (hyperphantasia).
Day-to-day observations of people across society demonstrate, and academic studies confirm, that people are able to complete tasks irrespective of whether their internal thoughts are represented through speech, imagery, or neither \citep{KEOGH2021237, anauralia}; though the lack of inner sight does impair the ability to recall visual characteristics \citep{Monzel2022, BAINBRIDGE2021159}.
Additionally, note that those possessing an inner monologue who speak multiple languages can have their inner monologue flip between languages depending on recent context.
These observations lead us to hypothesize that \textbf{conscious thoughts} (i.e. System~2 thinking) \textbf{are fundamentally abstract in nature}, but can be \textbf{projected to language and visual modalities} internally.

\subsection{What is System~2 reasoning?}

As a thought exercise, consider the task of solving this illustrative example from the Cognitive Reflection Test \citep{CognitiveReflectionTest}:
\begin{tcolorbox}
{A bat and a ball together cost \$1.10. The bat costs \$1 more than the ball. How much does the ball cost?}
\end{tcolorbox}

\textit{For illustrative purposes, please briefly solve the problem yourself before reading ahead.}
System~1 is responsible for the automatic response which immediately comes to mind on reading the problem: ten cents.
This answer is yielded in an involuntary manner, seemingly to all who hear the question for the first time.
However, by engaging System~2 we can verify whether the intuitive solution is correct, and reason about it.
By reflecting on the intuitive answer, one can observe that if this were the price of the ball, the total would be \$1.20, hence the answer is incorrect.
Since as the total price is the difference in price between the two objects (\$1) plus \textit{twice} the price of the ball, the answer is in fact 5 cents.

If we analyze it, it appears that the instinctive response stems from pattern matching---the problem looks at first glance like other problems comparing the quantity of two items, which we have solved in the past using the ``subtraction'' method, hence we instinctively try to apply it here.
Analogous mistakes can be seen in large language model (LLM) outputs, where they respond to a fake logic puzzle as if it were real \citep{Richardson2024}.

One way to conceptualize such reasoning is as a series of hypothesis generation and verification steps.
If the initial hypothesis fails the verification, we then come up with a new hypothesis conditioned on the new information generated during the verification process.
This process is repeated until a revised hypothesis satisfies the verification step.
Such a framework is similar to the Actor-Critic reinforcement learning algorithm \citep{ActorCritic}, with the actor analogous to the hypothesis generator and the critic as the verifier. 
This could be interpreted as alternating steps of System 1 (for hypothesis generation) and 2 (critique), however such a construction is contrary to prevalent psychological models of cognitive processes such as the dual process theory.


Alternatively, human reasoning can be conceptualized as a train of thought in a continuous stream of consciousness \citep{Potter2014, james2012principles}.
This framework is comparable to the chain-of-thought LLM prompting technique \citep{cot2022}, in which a model reaches its final output by focusing on a series of steps.

\subsection{Existing neural reasoning agents}

Previous work has found that by prompting LLMs with an in-context example of chain-of-thought reasoning, and asking it to think step-by-step for its own answer, models can be coerced into ``thinking'' step-by-step \citep{cot2022}.
Providing such a prompt changes the distribution of most likely initial next tokens to be steps towards the solution instead of an immediate answer.
By having the model attend to its own outputs as it progresses, it can build on its previous steps, eventually producing a final result that is more likely to be accurate \citep{cot2022, li2024chain}.
However, recent work has demonstrated the majority of the gains seen when using chain-of-thought prompting can be matched by prompting with a long series of task-independent filler tokens instead, suggesting the length of the sequence and the size of the compute graph is more important than the textual output \citep{Pfau2024}.
This implies the transformer can process data through unseen computations within the hidden layers of the network, unwitnessed in the chain-of-thought tokens that it outputs.
Such findings may be analogous to System~2 reasoning in humans, which we noted in \autoref{sec:bg:modes-of-thought} are primarily non-symbolic but can be projected to consciously observed language streams \citep{Hurlburt2013}, though such a hypothesis is challenging to investigate due to the difficulties of interpreting deep transformer representations \citep{rai2024practicalreviewmechanisticinterpretability}.

In the domain of closed-world games, tremendous gains were seen by applying deep reinforcement learning models that learn through self-play to optimize a value function \citep{AlphaGo, AlphaZero, Silver2018}.
In this case, the value network can be fit to predict the likelihood of each player winning the game from a given position.
The result of the game can be objectively determined by continuing to play the match and seeing who wins, providing a strong training signal to the value network.
And since the value network is able to fit this task, it is able to steer the actor model effectively.
Results are further improved by performing a Monte Carlo Markov chain tree search at inference time, guided by the model's predictions to prune the tree to a narrow range of feasible moves, to evaluate future game states and choose an optimal move.
Such searches are similar to the Tree-of-thoughts approach to improve chain-of-thoughts reasoning \citep{long2023largelanguagemodelguided, yao2023treethoughtsdeliberateproblem}.

Similarly, when deploying LLMs on mathematics problems, step-level verification of chain-of-thought has been shown to be effective training technique \citep{cobbe2021trainingverifierssolvemath, uesato2022solvingmathwordproblems, Lightman2024}.



\section{Future steps and potential pitfalls}

\subsection{Learning to reason}

Given the existing neural reasoning techniques, and their analogous relationship to human reasoning processes, we posit that networks already can learn to reason.

Multiple works have shown training LLMs to reason step-by-step is best achieved by step-level feedback \citep{zelikman2022star, Pfau2024, Lightman2024}.
One issue for training a reasoning model at scale is thus that there is a lack of large-scale reasoning datasets to train on in which humans have written out their train of thoughts explicitly.
However, such data can be acquired at modest scale (e.g. \citealp{hotpotqa}), and, by explicitly labelling which steps are valid and which are not, such data can be used to train a verifier that predicts whether individual logical reasoning steps are sound.
This verifier, similar to the rationalization evaluator used by \citet{zelikman2022star}, can then serve a similar role to the step-wise maths problem solver of \citet{Lightman2024}.
Using this, we can bootstrap more chain-of-thought data by tasking a pretrained LLM with chain-of-thought prompting to generate more reasoning data, and discarding outputs which contain steps which do not pass the verifier, similar to that used for maths problems by \citet{Lightman2024}.

Note that the verifier is an essential part of this pipeline, and it must be accurate in order for the iterative self-distillation to be effective.
But in any scenario where verification is easier than generation, the verifier (even if learnt and imperfect) can be deployed to iteratively refine and distill the generative model \citep{Christiano2018}.
An alternative bootstrap formulation would be to generate a large body of chain-of-thoughts data using chain-of-thought prompting applied on a large corpus of problems with known solutions.
We then train a verifier to, given a particular point in the chain-of-thought, classify whether the model will get the right answer.
This verifier model will serve a similar role to the value function in self-play RL systems \citep{Silver2018}, and we can fine-tune our model to generate its step of thoughts whilst trying to maximize the verifier's probability the problem will be solved.
Since such a system bears similarity to Q-learning and STaR bootstrap reasoning \citep{zelikman2022star}, it might be aptly given the name ``Q*''.
We note that other recent work has successfully applied reinforcement learning fine-tuning to pretrained LLMs, such as reinforcement with human feedback (RLHF) \citep{rlhf} or with harmlessness feedback \citep{rlaif}; and these methods can be improved by modifying the method to provide direct feedback \citep{dpo, drlaif}.
The implementation we propose would be a similar reinforcement learning fine-tuning stage, but with a objective focused on reasoning accuracy.

When such a model is half-way through its reasoning at deployment time, the steps that the model has already produced are provided back to it as in input to generate the next step.
However, the inaccuracy of the model creates a domain-shift between its training data and the data it sees during inference---during training teacher forcing means the model only sees accurate steps, but at deployment time the previous steps can be incorrect.
From a reinforcement learning perspective, such a training configuration is ``off-policy'', which can be corrected for by (re)training the model on its own outputs (on-policy), allowing it to learn to back-track when its inputs include incorrect steps \citep{kumar2024traininglanguagemodelsselfcorrect}.
We note that the importance of backtracking rises \textit{exponentially} with the number of steps in the reasoning process, since every reasoning step is an opportunity to make a mistake.

All the components for this solution for a reasoning agent framework seemingly already exist in the literature, and it is even possible such a model has already been trained recently \citep{o1}.

\subsection{Applicability}

LLMs trained only on textual data are unlikely to master reasoning about the real-world, since their observations of it are highly indirect.
When humans communicate with each other, they do so with a large body of common experiences merely from both being creatures raised and living in the real world.
This means that many things that are taken for granted remain unstated as they are assumed to be known by all parties in the discourse.

In order for foundation models to be able to reason efficiently about the world, we speculate they will need a world model that is built on sensory observations, not just text descriptions.
More recent foundation models have made progress in this direction \citep{visual-commonsense} by being multi-modal---processing both language and visual stimuli.
However, we posit that further gains will be made when using data which captures the richness of the real world through video data and (less abundantly) embodied sensorimotor data.
Video data has rich features about the world, enabling the network to construct its own intuitive physics, infer cause and effect \citep{VJEPA}.

\subsection{Scaling}

Will scaling laws continue to hold for chain-of-thought reasoning, or will such models hit scaling problems?

The ``bitter lesson'' of machine learning has been that gains from methods that can exploit generic compute scaling (e.g. larger and more flexible models, trained increasingly large datasets), in the long-run outperform gains from human-knowledge adjustments due to Moore's law \citep{sutton2019bitter}.
Thus we postulate that reasoning models will naturally also benefit from utilizing general methods rather than hand-tuned routines.
This is evidenced by recent work deploying LLMs on mathematical problems \citep{Snell2024}, which found that evaluation performance increases as the amount of inference compute increases.

However, one possible obstacle is the quadratic scaling 
of transformers with respect to their input sequence length due to their all-to-all attention.
Inefficient chain-of-thought reasoning will create excessively verbose thought-histories, greatly increasing the amount of compute required to reach the end of a chain-of-thought.
This poses a challenge to efficiently utilize compute when the model's inference steps are scaled up.
There have been various attempts to modify transformers to scale better \citep{sparse_transformers, performers, flashattention, flashattention2}.
Recently there have also been orthogonal efforts towards SOTA LLMs that are built using State Space Model (SSM) architectures \citep{s4, hyena, mamba, mamba2}.

More critically, as the number of entities to reason about grows, the number of potential interactions between the entities grows exponentially.
This has the potential to out-scale the computational resources available to train and deploy reasoning models.
However, we note that human working memory is limited to $7\pm2$ objects or chunks across a variety of tasks, where the number and size of chunks depends on the individual's familiarity with the items being held in memory \citep{Miller1956-uz}.
This implies that reasoning does not require all-to-all attention over objects in the thought history, rather it only requires a constant memory space.
The remaining challenges are (1)~items being held in memory must be appropriately compact; (2)~when only a limited number of items are retained in memory, the model must learn which memories to keep and which to drop.

With regards to compactness, this is a challenge for token-based models as typically the embedding space has the same granularity as the stimulus space.
Yet recent hierarchical models from the vision literature offer insights into how a hierarchical token-based model may look, in which the embedding space is more spatially compact than the stimulus representations \citep{swin, mvit, mvitv2, hiera}.

With regards to selecting memories to retain, recent work on memory-augmented transformers \citep{Bulatov2024} and on SSMs that can select and retain memories in their state-space \citep{mamba2} each provide research directions towards this goal, though there is still work to be done.
Even if memory selection remains challenging, less efficient reasoning models will be possible in the meantime.

\hider{
\subsection{Evaluation}

Current evaluations of LLMs typically do not take into account that longer responses require more compute, and more compute should lead to a better response.
We thus propose that inference evaluation should considered as the Pareto-front of both performance metric and compute resources expended.

Some work has already begun on constructing evaluation benchmarks for planning and reasoning systems such as PlanBench \citep{PlanBench}.
How to create 
} 

\subsection{Safety concerns}

As new capabilities are introduced to AI models, it is important to monitor these frontier models for potential safety risks \citep{phuong2024evaluatingfrontiermodelsdangerous}.
From an AI control perspective, ML agents which can reason and strategically plan present a much larger risk than passive models which merely predict things.
Like any ML model in deployment, there is a societal risk that the model's learnt biases from its training distribution will result in its behaviour diverging from human aspirations.

But more importantly, such a model raises the existential risk from AI models.
Models which can reason can use their abilities to plan and strategize, potentially over the long-term.
If allowed to act autonomously to achieve a goal, they may erroneously or surreptitiously plan with subgoals that involve taking control of resources they should not have access to, etc\footnote{We will not go into unnecessarily explicit details, as it is plausible this paper may be included in the training data of future LLMs.}.
To mitigate these concerns, it is important that training data be screened to ensure it does not contain instructions we would not wish an agent to take when deployed in the wild.

Another concern regards the scrutability of reasoning agents.
Current LLMs must always project their chain-of-thought reasoning steps to English, though there are concerns that their internal computation may not be fully reflected in their outputs \citep{Pfau2024, lyu-etal-2023-faithful, lanham2023measuringfaithfulnesschainofthoughtreasoning}.
From a gain of function perspective, it may be advantageous to train models that can reason in abstract concepts that do not directly correspond to tokens in the training corpus.
However, we are of the opinion that steps must always be taken to ensure that model reasoning is projected into a frame (be it language or imagery) in which it can be explicitly and as completely as possible communicated to humans.

\section{Conclusions}

We have discussed the literature surrounding the philosophy of human inner thought and reasoning, and the current neural network approaches to reasoning models.
The current networks have strong analogues to processes ascribed to human reasoning.
We thus argue they already achieve reasoning, though to limited degrees due to either their limited domains or lack of explicit training.

From this, we propose a pipeline which combines several existing techniques from the machine learning literature together as a candidate for how a reasoning agent could be explicitly trained to reason.
By expanding the breadth of training data to include richer, raw, temporal stimuli such as video, we anticipate the model can achieve a more capable world model to anchor its representations and better reason about the real world.
Thus we conclude that neural reasoning models are either already here, or if not they will be soon.

\begin{ack}
Many thanks to David Emerson, Iulia Eyriay, Kevin Kasa, Kristen Menou, and Michael Zhang for insightful discussions and feedback, and to Philip from AI Explained for providing the initial inspiration \citep{aiexplained}.

Resources used in preparing this work were provided, in part, by the Province of Ontario, the Government of Canada through CIFAR, and \hreffoot{https://vectorinstitute.ai/partnerships/current-partners/}{companies sponsoring} the Vector Institute.
\end{ack}


\bibliography{main}
\bibliographystyle{icml2024}

\end{document}